\documentclass{article}
\usepackage{ijcai23}

\usepackage{times}
\usepackage{soul}
\usepackage{url}
\usepackage[hidelinks]{hyperref}
\usepackage[utf8]{inputenc}
\usepackage[small]{caption}
\usepackage{graphicx}
\usepackage{amsmath}
\usepackage{amsthm}
\usepackage{enumitem}
\usepackage{booktabs}
\usepackage{mathtools}
\usepackage{natbib}

\usepackage{setspace}
\usepackage{multirow}
\usepackage{bm,bbm}
\usepackage{enumitem}
\usepackage{amsmath}
\usepackage{amssymb}
\usepackage{amsthm}
\usepackage{mathtools}
\usepackage{amsfonts}
\usepackage{booktabs}       
\usepackage{nicefrac}  
\usepackage{amsthm}

\usepackage[switch]{lineno}
\usepackage[ruled,noresetcount,vlined,linesnumbered]{algorithm2e}
\usepackage{xcolor}

\DeclareMathOperator*{\argmax}{argmax}

\newtheorem{theorem}{Theorem}

\title{Differentiable Model Selection for Ensemble Learning}

\author{James Kotary$^{1}$  
\and Vincenzo Di Vito$^{1}$ 
\And Ferdinando Fioretto$^{2}$ \\
\affiliations
  $^1$
  Syracuse University\\
  $^2$
  University of Virginia\\
\emails
\{jkotary, vdivitof\}@syr.edu,
nandofioretto@gmail.com
}

\begin{document}

\maketitle

\begin{abstract}
\emph{Model selection} is a strategy aimed at creating accurate and robust models. A key challenge in designing these algorithms is identifying the optimal model for classifying any particular input sample. This paper addresses this challenge and proposes a novel framework for \emph{differentiable model selection} integrating machine learning and combinatorial optimization.
The framework is tailored for ensemble learning, a strategy that combines the outputs of individually pre-trained models, and learns to select appropriate ensemble members for a particular input sample by transforming the ensemble learning task into a differentiable selection program trained end-to-end within the ensemble learning model. Tested on various tasks, the proposed framework demonstrates its versatility and effectiveness, outperforming conventional and advanced consensus rules across a variety of settings and learning tasks.
\end{abstract}

\section{Introduction}
\label{sec:Intro}

Model selection involves the process of identifying the most suitable model from a set of candidates for a given learning task. The chosen model should ideally generalize well to unseen data, with the complexity of the model playing a crucial role in this selection process. However, striking a balance between underfitting and overfitting is a significant challenge.

A variety of techniques have been presented in the machine learning literature to address this issue. Of particular relevance, {\em ensemble learning} \citep{witten2005practical} is a meta-algorithm that 
combines the outputs of individually pre-trained models, known as \emph{base learners}, to improve overall performance.
Despite being trained to perform the same task, these base learners may exhibit error diversity, meaning they fail on different samples, and their accuracy profiles complement each other across an overall distribution of test samples. The potential effectiveness of an ensemble model strongly depends on the correlation between the base learners' errors across input samples and their accuracy; those with higher accuracy and error diversity have a higher potential for improved ensemble accuracy \citep{mienye2022survey}. 

However, the task of identifying the optimal ensemble of models for classifying any particular input sample is nontrivial. Traditional approaches often aggregate predictions across all base learners of an ensemble, aiming to make predictions more robust to the error of individual base learners. While these techniques could be enhanced by selectively applying them to a subset of base learners known to be more reliable on certain inputs, the design of algorithms that effectively select and combine the base learners' individual predictions remains a complex endeavor. Many consensus rule-based methods apply aggregation schemes that combine or exclude base learners' predictions based on static rules, thereby missing an opportunity to inform the ensemble selection based on a particular input's features.

Recently, the concept of differentiable model selection has emerged, aiming to incorporate the model selection process into the training process itself \citep{dona2021differentiable,sheth2020differentiable,fu2016drmad}. This approach leverages gradient-based methods to optimize model selection, proving particularly beneficial in areas like neural architecture search. The motivation behind differentiable model selection lies in the potential to automate and optimize the model selection process, thereby leading to superior models and more efficient selection procedures.
Despite its promises, however, it remains non-trivial how to design effective differentiable model selection strategies and the use of gradient-based methods alone further enhances the risk of converging to local optima which can lead to suboptimal model selection.

In light of these challenges, this paper proposes a novel framework for differentiable model selection specifically tailored for ensemble learning. This framework integrates machine learning and combinatorial optimization to learn the selection of ensemble members by modeling the selection process as an optimization problem leading to optimal selections within the prescribed context. 

\paragraph{\bf Contributions.}
In more detail, this paper makes the following contributions: {\bf (1)} It proposes \emph{end-to-end Combinatorial Ensemble Learning (e2e-CEL)}, a novel ensemble learning framework that exploits an integration of ML and combinatorial optimization to compile specialized consensus rule \emph{for a particular input sample}. 
{\bf (2)} It shows how to cast the selection and aggregation of ensemble base learner predictions as a differentiable optimization problem, which is parameterized by a deep neural network and trained end-to-end within the ensemble learning task. 
{\bf (3)} An analysis of challenging learning tasks demonstrates the strengths of this idea: e2e-CEL outperforms models that attempt to select individual ensemble members, such as the optimal weighted combination of the individual ensemble members' predictions as well as conventional consensus rules, implying a much higher ability to leverage error diversity. 

These results demonstrate the integration of constrained optimization and learning to be a key enabler 
to enhance the effectiveness of model selection in machine learning tasks. 

\section{Related Work}

Ensemble learning involves two steps: training individual base learners and combining their outputs for accurate predictions. The composition of an ensemble from base learners with complementary error profiles is commonly done through \emph{bagging} (randomly creating datasets for training each member) and \emph{boosting} (adaptively creating datasets based on error distributions to increase error diversity). 
A survey of training individual base learners can be found in \cite{mienye2022survey}.
The second step is typically handled by classical aggregation rules over the predictions or activation values of ensemble members, such as majority or plurality voting. Some works have also attempted to mathematically model more effective aggregation rules, such as the \emph{Super Learner} algorithm \citep{ju2018relative} which forms a weighted combination of base learner models that maximizes accuracy over a validation set. This algorithm has been proven to be asymptotically optimal for combining ensemble members predictions.

{\em This paper addresses the latter, challenging, aspect of ensemble modeling}: optimizing the aggregation of predictions from individual ensemble base learners. 
The proposed e2e-CEL approach aims to learn aggregation rules adaptively at the level of individual input samples, rather than a single rule for all samples. 
While heuristic-based selection rules to derive input-dependent ensembles are not new to the literature, to the best of our knowledge, this is the first proposal of a method that learns such conditional rules in an end-to-end manner.
A discussion on additional work is deferred to Appendix A.

\section{Setting and Goals}

\label{sec:SettingAndGoals}

The paper considers \emph{ensembles} as a collection of $n$ models or \emph{base learners} represented by functions $f_i, 1 \leq i \leq n$, trained independently on separate (but possibly overlapping) datasets ($\mathcal{X}_i$,$\mathcal{Y}_i$), all on the same intended \emph{classification} task.  On every task studied, it assumed that ($\mathcal{X}_i$,$\mathcal{Y}_i$) are given, along with a prescription for training each base learners, so that $f_i$ are assumed to be pre-configured.
This setting is common in federated analytic contexts, where base learners are often trained on diverse datasets with skewed distributions \citep{federatedLearning}, and in ML services, where providers offer a range of pre-trained models with different  architectural and hyper-parametrization choices \citep{MLAS}. 

Let $n \!\in\! \mathbb{N}$ be the number of base learners, $c \!\in\! \mathbb{N}$ the number of classes and $d \!\in\! \mathbb{N}$ the input feature size. Given a sample $z \!\in\! \mathbb{R}^d$, each base learner $f_j \colon \mathbb{R}^d \!\to\! \mathbb{R}^c$ computes $f_j(z) = \hat{y}_{j}$. For the classification tasks considered in this paper, each $\hat{y}_j$ is the direct output of a  \textit{softmax} function $\mathbb{R}^c \!\rightarrow\! \mathbb{R}^c$,
\begin{equation}
    \label{eq:softmax}
    \texttt{softmax}(c)_i = \frac{e^{c_i}}{\sum_{k=1}^c e^{c_k}}.
\end{equation}

Explicitly, each classifier $f_i(\phi_i, x)$ is trained with respect to its parameters $\phi_i$ to minimize a classification loss $\mathcal{L}$ as 
\begin{equation}
    \min_{\phi_i} \mathbb{E}_{(x,y)\sim (\mathcal{X}_i,\mathcal{Y}_i)} \left[ \mathcal{L} (f_{i}(\phi_i,x),y)  \right].
\end{equation}
The goal is then to combine the base learners into an \emph{ensemble}, whose aggregated classifier $g$ performs the same task, but with greater overall accuracy on a \emph{master dataset}
($\mathcal{X}$,$\mathcal{Y}$), where $\mathcal{X}_i \subset \mathcal{X}$
and $\mathcal{Y}_i \subset \mathcal{Y}$ for all $i$ with $0 \leq i \leq n$:
\begin{equation}
    \min_{\theta} \mathbb{E}_{(x,y)\sim (\mathcal{X},\mathcal{Y})} \left[ \mathcal{L} (g(\theta,x),y)  \right].
\end{equation}
As is typical in ensemble learning, the base learners may be 
trained in a way that increases test-error diversity among $f_i$ on 
$\mathcal{X}$ --- see Section \ref{sec:Experiments}. In each dataset there
is an implied train/test/validation split, so that evaluation of a 
trained model is always performed on its test portion. Where this 
distinction is needed, the symbols $\mathcal{X}_{\texttt{train}}$, 
$\mathcal{X}_{\texttt{valid}}$, $\mathcal{X}_{\texttt{test}}$ are used. 
A list of symbols used in the paper to describe various aspects of the computation, along with their meanings is provided in \citep{kotary2022endtoend}, Table 4.

\begin{figure*}
    \centering
    \includegraphics[width=0.8\linewidth]{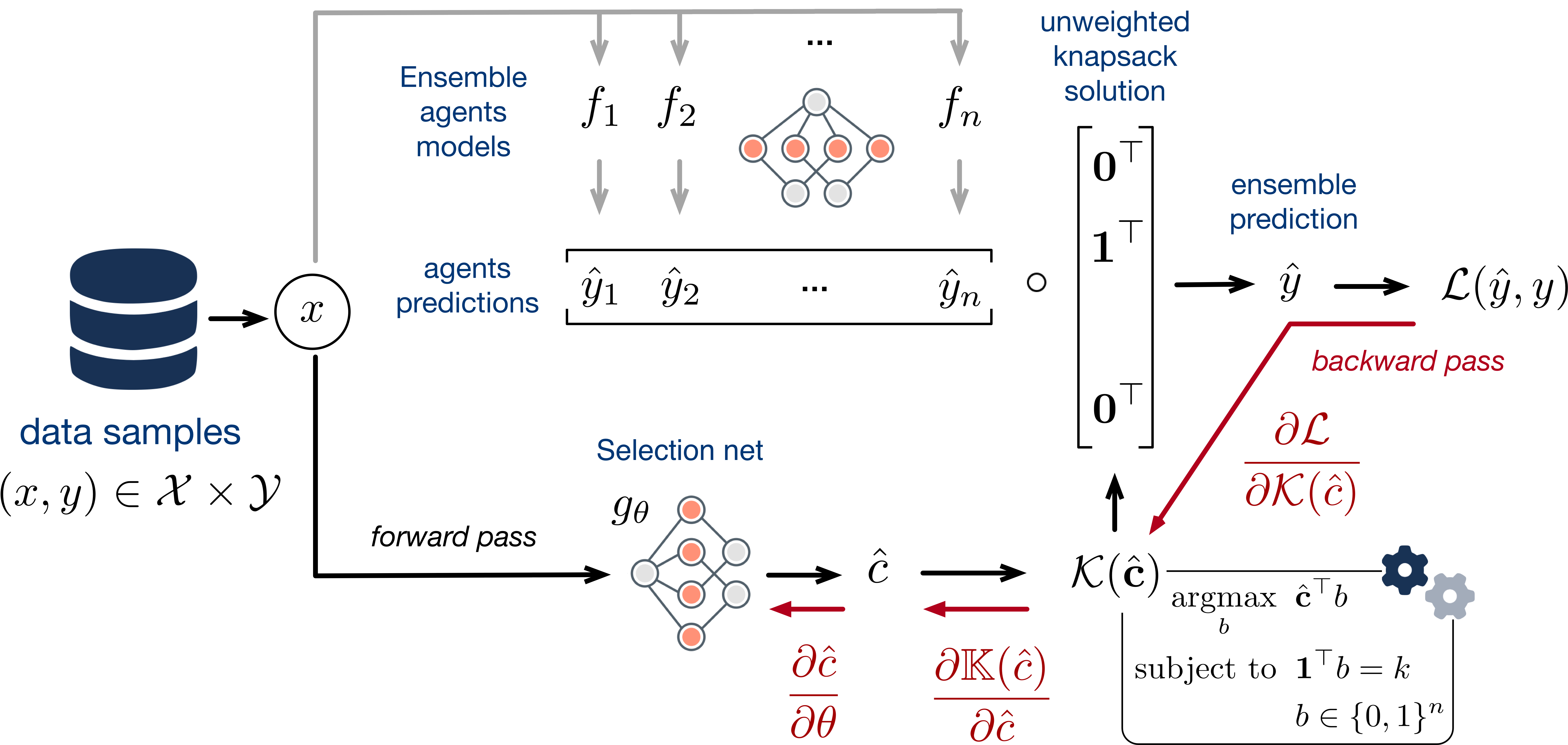}
    \caption{
    End-to-end Ensemble Learning scheme:
    Black and red arrows illustrate forward and backward operations, 
    respectively. 
    \label{fig:e2e_scheme}}
\end{figure*}

\section{End-to-end Combinatorial Ensemble Learning}
\label{sec:e2eMEL}

Ideally, given a pretrained ensemble $f_i, 1 \leq i \leq n$ and a 
sample $z \in \mathcal{X}$, one would select from the ensemble a 
classifier which is known to produce an accurate class prediction for 
$z$. However, a performance assessment for each base learners' predictions 
is not available at test time. Thus, conventional ensemble learning 
schemes resort to selection criteria such as \emph{plurality voting} (see Section \ref{sec:Experiments} for a description of the aggregation rules here used as a benchmark).

The end-to-end learning scheme in this work is based on the idea that a more accurate ensemble prediction can be made by using predictions based on $z$, and that selecting a well-chosen subset of the ensemble, rather than the entire ensemble, can provide more reliable results than a single base learner. The size of the subset, $k$, is treated as a hyperparameter. While it may seem logical to only choose the best predicted base learner for a given input sample (setting $k=1$), it has been consistently observed in the experiments that the optimal performance is achieved for $1 < k < n.$

The proposed mechanism casts the sub-ensemble selection as an optimization program that is end-to-end differentiable and can thus be integrated with a learning model $g_{\theta}$ to select a reliable subset of the ensemble base learners to combine for  predictions. An \emph{end-to-end Combinatorial Ensemble Learning} (e2e-CEL), or simply, \emph{smart ensemble}, consists of an ensemble of base learners  along with a module that is trained by e2e-CEL to select the \emph{sub-ensemble} of size $k$, which produces the most accurate combined prediction for a given input. The model $g$ is called the \emph{selection net}, and the end-to-end ensemble model is trained by optimizing its parameters $\theta$.

\paragraph{E2e-CEL overview.} E2e-CEL is composed of three main steps:
\begin{enumerate}[labelsep=2pt,itemsep=2pt,parsep=0pt,topsep=2pt]
    \item Predict a vector of scores $g_{\theta}(z) = \hat{c}$, estimating the prediction accuracy for each base learner on sample $z$.
    \item Identify the base learner indices $\mathcal{E} \subset [n]$ which correspond the the top $k$ predicted scores.
    \item Collect the predictions of the selected sub-ensemble $f_j(z)$ and perform an approximate majority voting scheme over those predictions to determine the $z$'s class.
\end{enumerate}

By training on the master set $\mathcal{X}_{\texttt{train}}$, 
the smart ensemble learns to make better predictions by
virtue of learning to select better sub-ensembles to vote on 
its input samples. However, note that subset selection and 
plurality voting are discrete operations, and in plain form do not offer useful gradients for backpropagation and learning. The next sections discuss further details of the e2e-CEL framework, including differentiable approximations for each step of the overall model.

Figure \ref{fig:e2e_scheme} illustrates the e2e-CEL model and its training process in terms of its component operations. Backpropagation is shown with red arrows, and it only applies to the operations downstream from the selection net $g$, since the e2e-CEL is parameterized by the parameters of $g$ alone.

\subsection{Differentiable Model Selection}
\label{sec:diff_model_select}

The e2e-CEL system is based on learning to select $k<n$ predictions from the master ensemble, given a set of input features. This can be done by way of a structured prediction of binary values, which are then used to mask the individual base learner predictions. 

    Consider the unweighted knapsack problem 
\begin{subequations}
\label{eq:knapsack}
    \begin{align}
    \mathcal{K}(\hat{c}) = 
    \underset{b}{\text{argmax }} &\;\;
    \hat{c}^\top b\label{eq:optProb}\\
    \text{subject to }&\;\;
    \mathbf{1}^\top b = k,\label{eq:constraint1}\\
    &\;\; b \in \{0,1\}^n, \label{eq:int_constraint}
    \end{align}
\end{subequations}
which can be viewed as a selection problem whose optimal solution assigns the value $1$ to the elements of $b$ associated to the top $k$ values of $\hat{c}$. Relaxing constraint (\ref{eq:int_constraint}) to $0 \leq b \leq 1$ results in an equivalent linear program (LP) with discrete optimal solutions  $b \in \{ 0,1 \}^n$, despite being both convex and composed of continuous functions. This useful property holds for any LP with totally unimodular constraints and integer right-side coefficients \citep{bazaraa2008linear}. 

This optimization problem can be viewed as a mapping from $\hat{c}$ to a binary vector indicating its top $k$ values, and represents thus a natural candidate for selection of the optimal sub-ensemble of size $k$ given the individual base learners' predicted scores, seen as $\hat{c}$.
However, the outputs of Problem \eqref{eq:knapsack} define a piecewise constant function, $\mathcal{K}(\hat{c})$, which does not admit readily informative gradients, posing challenges to differentiability. 
For integration into the end-to-end learning system, the function $\mathcal{K}(\hat{c})$ must provide informative gradients with respect to $\hat{c}$. 
In this work, this challenge is overcome by smoothing $\mathcal{K}(\hat{c})$ based on perturbing $\hat{c}$ with random noise.

As observed by \cite{berthet2020learning}, any continuous, convex linear programming problem can be used to define a \emph{differentiable perturbed optimizer}, which yields approximately the same solutions but is differentiable with respect to its linear objective coefficients. Given a random noise variable $Z$ with probability density  $p(z) \propto \exp{(-v(z))}$ where $v$ is a twice differentiable function,
\begin{equation}
    \label{eqn:DPOfwd}
    \mathbb{K}(\hat{c}) = \mathbb{E}_{z \sim Z}\;\; \left[ \mathcal{K}(\hat{c} + \epsilon z)  \right],
\end{equation}
is a differentiable perturbed optimizer associated to $\mathcal{K}$. The temperature parameter $\epsilon > 0$ controls the sensitivity of its gradients (or properly, Jacobian matrix), which can itself be represented  by the expected value \citep{abernethy2016perturbation}:
\begin{equation}
    \label{eqn:DPObwd}
    \frac{\partial \mathbb{K}(\hat{c})}{\partial \hat{c}} = \mathbb{E}_{z \sim Z}\;\; \left[ \mathcal{K}(\hat{c} + \epsilon z) \;\; v'(z)^\top  \right].
\end{equation}

In this work, $Z$ is modeled as a standard Normal random variable. While these expected values are analytically intractable (due to the constrained $\argmax$ operator  within the knapsack problem $\mathcal{K}$), they can be estimated to arbitrary precision by sampling in Monte Carlo fashion. This procedure is a generalization of the Gumbel Max Trick \citep{gumbel1954statistical}.  

Note that simulating Equations (\ref{eqn:DPOfwd}) and (\ref{eqn:DPObwd}) requires solving Problem \eqref{eq:knapsack} for potentially many values of $z$. However, although the theory of perturbed optimizers requires the underlying problem to be a linear program, only a blackbox implementation is required to produce $\mathcal{K}(\hat{c})$ \citep{berthet2020learning}, allowing for an efficient algorithm to be used in place of a (more costly) LP solver.  The complexity of evaluating the differentiable perturbed optimizer $\mathbb{K}(\hat{c})$  is discussed next.
\begin{theorem}
The total computation required for solving Problem \eqref{eq:knapsack} is  $\mathcal{O}( n \log k)$, where $n$ and $k$ are, respectively, the ensemble and sub-ensembles sizes.
\end{theorem}
\begin{proof}
This result relies on the observation that $\mathcal{K}(\hat{c})$ can be computed efficiently by identifying the top $k$ values of $\hat{c}$ in $\mathcal{O}(n \log k)$ time using a divide-and-conquer technique. See, for example, \citep{cormen2022introduction}. 
\end{proof}

Generating $m$ such solutions for gradient estimation then requires $\mathcal{O}(m\,n \log k)$ operations. Note, however, that these can be performed in parallel across samples, allowing for sufficient noise samples to be generated for computing accurate gradients, especially when GPU computing is available. 

\smallskip
For clarity, note also that the function $\mathcal{K}$, as a linear program mapping, has a discrete output space since any linear program takes its optimal solution at a vertex of its feasible region \citep{bazaraa2008linear}, which are finite in number. As such, it is a piecewise constant function and is differentiable except on a set of measure zero \citep{folland1999real}. However, \mbox{$\nicefrac{\partial \mathcal{K}}{\partial \hat{c}} = 0$} everywhere it is defined, so the derivatives lack useful information for  gradient descent \citep{wilder2019melding}. While $\nicefrac{\partial \mathbb{K}}{\partial \hat{c}}$ is not the true derivative of $\mathcal{K}$ at $\hat{c}$, it supplies useful information about its direction of descent. 

In practice, the forward optimization pass is modeled as $\mathcal{K}(\hat{c})$, and the backward pass is modeled as $\nicefrac{\partial \mathbb{K}(\hat{c})}{\partial \hat{c}}$. This allows further downstream operations, and their derivatives, to be evaluated at $\mathcal{K}(\hat{c})$ without approximation, which improves training and test performance. These forward and backward passes together are henceforth referred to as the \emph{Knapsack Layer}. Its explicit backward pass is computed as
\begin{equation}
    \label{eqn:DPObwd_approx}
    \frac{\partial \mathbb{K}(\hat{c})}{\partial \hat{c}} \approx \frac{1}{m} \sum_{i = 1}^m\;\; \left[ \mathcal{K}(\hat{c} + \epsilon z_i) \;\; v'(z_i)^\top  \right],
\end{equation}
where $z_i \sim \mathcal{N}(0,1)^n$ are $m$ independent samples each drawn from a standard normal distribution. 

\subsection{Combining Predictions}

Denote as $P \in \mathbb{R}^{{c}\times{n}}$ the matrix whose $j^\textrm{th}$ column is the softmax vector $\hat{y}_j$ of base learner $j$,
\begin{equation}
P=
\begin{pmatrix}
    \hat{y}_1 & \hat{y}_2 \dots \hat{y}_n.\\
\end{pmatrix}
\end{equation}

For the purpose of combining the ensemble base learner predictions, $\mathcal{K}(\hat{c})$ is treated as a binary masking vector $b \in \{0,1\}^n$, which selects the subset of base learners for making a prediction. Denote as $B \in \{0,1\}^{c \times n}$ the matrix whose $i^{\textrm{th}}$ column is $B_i = \overrightarrow{\bf{1}} b_i $; i.e.,
$$
B =\left[ \begin{matrix}
        b_1 &
        \ldots &
        b_n
       \end{matrix}  \right]^\top.
$$
\noindent This matrix is used to mask the base learner models' softmax predictions $P$  by element-wise multiplication. Next, define
\begin{align}
    P_k &= B \circ P \notag\\ 
    & =\left[ 
    \begin{matrix} b_1 & \ldots & b_n \end{matrix}  
    \right]^\top \circ
    \left[ 
    \begin{matrix} \hat{y}_1 & \ldots & \hat{y}_n \end{matrix}  
    \right].
\end{align}

Doing so allows compute the sum of predictions over the selected sub-ensemble $\mathcal{E}$, but in a way that is automatically differentiable, that is:
\begin{equation}
    \hat{v} \coloneqq \sum_{i \in \mathcal{E}} \hat{y}_i = \sum_{i=1}^n P_k^{(i)}.
\end{equation}

\noindent The e2e-CEL prediction comes from applying softmax to this sum:
\begin{align}
    \hat{y} &= \texttt{softmax}(\hat{v}) 
    =
        \label{eqn:phat}
        \texttt{softmax}\left(\sum_{i=1}^n P_k^{(i)}\right),
\end{align}
viewing the softmax as a smooth approximation to the argmax function as represented with one-hot binary vectors. This function is interpreted as a smoothed majority voting to determine a class prediction: given one-hot binary class indicators $h_i$, the majority vote is equal to $\argmax (\sum_i h_i)$. An illustration of the process is given in Figure \ref{fig:e2e_scheme}. 

At test time, class predictions are calculated as 
\begin{equation}
    \argmax_{1 \leq i \leq c} \;\; \hat{y}_i(x).
\end{equation}

Combining predictions in this way allows for an approximated majority voting over a selected sub-ensemble, but in a differentiable way so that selection net parameters $\theta$ can be directly trained to produce selections that minimize the classification task loss, as detailed in the next section.

\subsection{Learning Selections}

The smart ensemble mechanism learns accurate class predictions by learning to select better subensembles to vote on its input samples. In turn, this is done by predicting better coefficients $\hat{c}$ which parameterize the Knapsack Layer. 

The task of predicting $\hat{c}$ based on input $z$ is itself learned by the \emph{selection net}, a neural network model $g$ so that $\hat{c} = g(z)$. Since $g$ acts on the same input samples as each $f_i$, it should be capable of processing inputs from $z$ at least as well as the base learners' models; in Section \ref{sec:Experiments}, the selection net in each experiment uses the same CNN architecture as that of the base learner models. Its predicted values are viewed as scores over the ensemble members, rather than over the possible classes. High scores correspond to base learners which are well-qualified to vote on the sample in question. 

In practice, the selection net's predictions $\hat{c}$ are normalized before input to the mapping $\mathcal{K}$: 
\begin{equation}
    \hat{c} \leftarrow \frac{\hat{c}}{{\lVert}{\hat{c}}{\rVert}}_2.
\end{equation}
This has the effect of standardizing the magnitude of the linear objective term in \eqref{eq:optProb}, and tends to  improve training. Since scaling the objective of an optimization problem has no effect on its solution, this is equivalent to standardizing the relative magnitudes of the linear objective and random noise perturbations in Equations \eqref{eqn:DPOfwd} and \eqref{eqn:DPObwd}, preventing $\epsilon$ from being effectively absorbed into the predicted $\hat{c}$.

For training input $x$, let $\hat{y}_\theta(x)$ represent the associated e2e-CEL prediction given the selection net parameters $\theta$. During training, the model minimizes the classification loss between these predictions and the ground-truth labels:
\begin{equation}
    \min_{\theta} \mathbb{E}_{(x,y)\sim (\mathcal{X},\mathcal{Y})} \left[ \mathcal{L} (\hat{y}_{\theta}(x),y)  \right].
\end{equation}

Generally, the loss function $\mathcal{L}$ is chosen to be the same as the loss used to train the base learner models, as the base learners are trained to perform the same classification task.

\subsection{e2e-CEL Algorithm Details}
Algorithm \ref{alg:learning} summarizes the e2e-CEL procedure for training a selection net. Note that only the parameters of the selection net are optimized in training, and so only its downstream computations are backpropagated. This is done by the standard automatic differentiation employed in machine learning libraries \citep{NEURIPS2019_9015}, except in the case of the Knapsack Layer, whose gradient transformation is analytically specified by Equation \eqref{eqn:DPObwd_approx}.

For clarity, Algorithm \ref{alg:learning} is written in terms of operations that apply to a single input sample. In practice, however, minibatch gradient descent is used. 
Each pass of the training begins evaluating the base learner models (line 4) and sampling standard Normal noise vectors (line 5). The selection net predicts from input features $x$ a vector of base learner scores $g_\theta(x)$, which defines an unweighted knapsack problem $\mathcal{K}(g_\theta(x))$ that is solved to produce the binary mask $b$ (line 6). 

Masking is applied to the base learner predictions before being summed and softmaxed for a final ensemble prediction $\hat{y}$ (line 8). The classification loss $\mathcal{L}$ is evaluated with respect to the label $y$ and backpropagated in 3 steps: \textbf{(1)} The gradient $\frac{\partial \mathcal{L}}{\partial b}$ is 
computed by automatic differentiation backpropagated to the Knapsack Layer's output (line 9). \textbf{(2)} The chain rule factor $\frac{\partial b}{\partial \hat{c}}$ is analytically computed by the methodology of Section \ref{sec:diff_model_select} (line 10). \textbf{(3)} The remaining chain rule factor $\frac{\partial \hat{c}}{\partial \theta}$ is computed by automatic differentiation (line 11). Note that as each chain rule factor is computed, it is also applied toward computing $\frac{\partial \mathcal{L}}{\partial \theta}$ (line 12). Finally, a SGD step \citep{ruder2016overview} or one of its variants (\citep{diederik2014adam}, \citep{zeiler2012adadelta}) is applied to update $\theta$ (line 13). 

The next section evaluates the accuracy of ensemble models trained with this algorithm, on classification tasks using deep neural networks.

\begin{algorithm}[!tb]
  \caption{Training the Selection Net}
  \label{alg:learning}
  \setcounter{AlgoLine}{0}
  \SetKwInOut{Input}{input}
  \SetKwInOut{Output}{output}
  \SetKwRepeat{Do}{do}{while}
  \Input{$\mathcal{X}, \mathcal{Y}, \alpha, k, m, epsilon$\!\!\!\!\!\!\!\!\!\!}
  \For{epoch $k = 0, 1, \ldots$} 
  { 
    \ForEach{$(x,y) \!\gets\! (\mathcal{X}, \mathcal{Y})$ }
    {
        $\hat{y}_i  \!\gets\! f_i(x) \qquad\;\; \forall\; 1 \leq i \leq n$\\
        $z_i \sim \mathcal{N}(0,1)^n \quad \forall\; 1 \leq i \leq m$\\
    
        $\left( b, \hat{c} \right) 
            \gets   
            \left( \mathcal{K}(g_{\theta}(x)), 
                    \frac{g_{\theta}(x)}{\|g_{\theta}(x)\|_2} \right)$\\
                    
        $P_k \gets \left[ b, \ldots, b \right]^\top\!\!
                   \circ
                   \left[ \hat{y}_1, \ldots, \hat{y}_n \right]
         $\\
        $\hat{p} \gets \texttt{softmax}(\sum_{i=1}^n P_k^{(i)}) $\\
        $\nicefrac{\partial \mathcal{L}(\hat{p},y)}{\partial b} \gets \texttt{autodiff}$\\
        $\nicefrac{\partial b}{\partial \hat{c}} \gets \frac{1}{m} \sum_{i = 1}^m\;\; \left[ \mathcal{K}(\hat{c} + \epsilon z_i) \;\; v'(z_i)^\top  \right]$\\
        $\nicefrac{\partial \hat{c}}{\partial \theta} \gets \texttt{autodiff}$\\
        $\nicefrac{\partial \mathcal{L}(\hat{p},y)}{\partial \theta} \gets 
        \frac{\partial \mathcal{L}(\hat{p},y)}{\partial b} \cdot \frac{\partial b}{\partial \hat{c}} \cdot \frac{\partial \hat{c}}{\partial \theta}$\\
        $\theta \gets \theta - \alpha \frac{\partial \mathcal{L}(\hat{p},y)}{\partial \theta}$
     }
  }
\end{algorithm}

\section{e2e-CEL Evaluation}
\label{sec:Experiments}

The e2e-CEL training is evaluated on several vision classification tasks: {digit classification} on {\sl MNIST dataset} \citep{mnistDataset}, age-range estimation on {\sl UTKFace dataset} \citep{utkfaceDataset}, image classification on {\sl CIFAR10 dataset} \citep{cifarDataset}, and emotion detection on {\sl FER2013} dataset \citep{ferDataset}. 

Being an optimized aggregation rule, e2e-CEL is compared with state-of-the-art Super Learner algorithm \citep{ju2018relative} and the following widely adopted baseline aggregation rules when paired with a pre-trained ensemble :  
\begin{itemize}[leftmargin=*, labelsep=2pt,itemsep=2pt,parsep=0pt,topsep=2pt]
    \item \textit{Super Learner}: a fully connected neural network that, given the base learners' predictions, learns the optimal weighted combinations specialized for any input sample.
    
    \item \textit{Unweighted Average}: averages all the base learners' softmax predictions and then compute the index of the corresponding highest label score as the final prediction.   
    
    \item \textit{Plurality Voting}: makes a discrete class prediction from each base learner and then returns the most-predicted class. 

    \item \textit{Random Selection}: randomly selects a size-$k$ sub-en\-sem\-ble of base learners for making prediction and then applies the unweighted average rule to the selected base learners' soft predictions.
\end{itemize}

\paragraph{Experimental settings.} As described in Section \ref{sec:Intro} and in Appendix A, ensemble learning schemes are most effective when base learner models are accurate and have high error diversity. In this work, base learners are deliberately trained to have high error diversity with respect to input samples belonging to different classes. This is done by composing for each base learner model $f_i \; (1 \leq i \leq n)$ a training set $\mathcal{X}_i$ in which a subset of classes is over-represented, resulting base learners that specialize in identifying those classes. The exact class composition of each dataset $\mathcal{X}_i$ depends on the particular classification task and on the base learner's intended specialization. 

For each task, each base learners is designed to be specialized for recognizing either one or two particular classes. To this end, the training set of each base learner is partitioned to have a majority of samples belonging to a particular class, while the remaining part of the training dataset is uniformly distributed across all other classes by random sampling. Specifically, to compose the \emph{smart ensemble} for each task, a single base learner is trained to specialize on each possible class, and on each pair of classes (e.g., digits $1$ and $2$ in MNIST). When $c$ is the number of classes, the experimental smart ensemble then consists of $c + {c \choose 2}$ total base learners. 
Training a specialized base learner in this way generally leads to high accuracy over its specialty classes, but low accuracy over all other classes. Therefore in this experimental setup, no single base learner is capable of achieving high overall accuracy on the master test set $\mathcal{X}_{\texttt{test}}$. This feature is also recurrent in federated analytic models \citep{federatedLearning}.

Table \ref{tab:specialized_base learner_acc} shows the average accuracy of individual base learner models on their specialty classes and their non-specialty classes; reported, respectively as \emph{specialized accuracy} and \emph{complementary accuracy}. The reported \emph{overall} accuracy is measured over the entire master test set $\mathcal{X}_{\texttt{test}}$. This sets the stage for demonstrating the ability of e2e-CEL training to compose a classifier that substantially outperforms its base learner models on $\mathcal{X}_{\texttt{test}}$ by adaptively selecting sub-ensembles based on input features; see Section \ref{sec:Results}.

Note that, in each experiment, the base learner models' architecture design, hyperparameter selection, and training methods have not been chosen to fully optimize classification accuracy, which is not the direct goal of this work. Instead the base learners have been trained to maximize error diversity, and demonstrate the ability of e2e-CEL to leverage error diversity and compose highly accurate ensemble models from far less accurate base learner models, in a way that is not shared by conventional aggregation rules. Note also that improving base learner model accuracies would, of course, tend to improve the accuracy of the resulting ensemble classifiers. In each case, throughout this section, the e2e-CEL selection net is given the same CNN architecture as the base learner models which form its ensemble.

\begin{table}[tb]
\centering
\resizebox{0.85\linewidth}{!}
{
    \begin{tabular}{r ccc}
      \toprule
        & \multicolumn{3}{c}{\textbf{Accuracy (\%)}}\\
         \textbf{Dataset} & \textbf{Specialized}& 
         \textbf{Complimentary} & \textbf{Overall}  \\
      \midrule
      {\textbf{MNIST}} & 97.5 & 86.8 & 89.6 \\ 
      {\textbf{UTKFACE}} & 93.2 & 25.2 & 51.2 \\ 
      {\textbf{FER2013}} & 79.4 & 38.1 & 47.8 \\ 
      {\textbf{CIFAR10}} & 76.3 & 24.8 & 31.1 \\ 
      \bottomrule
    \end{tabular}
}
\caption{Specialized base learner model test accuracy}
\label{tab:specialized_base learner_acc}
\end{table} 

\subsection{Datasets and Settings}
For each task, the base learners are trained to specialize in classifying one or two particular classes, which allows the selection program to leverage their error diversity. Additional details about the base learners' models and the dataset split can be found in  Appendix B.

\paragraph{Digit classification.}
MNIST is a 28x28 pixel greyscale images of handwritten digits dataset, which contains 60000 images for training and 10000 images for testing.
The ensemble consists of $55$ base learners, $10$ of which specialize on $1$ class  and $ {10 \choose 2} = 45$ of which specialize on $2$ classes.

\paragraph{Image classification.}
CIFAR10 is a 32x32 pixel color images dataset in 10 classes: airplanes, cars, birds, cats, deer, dogs, frogs, horses, ships, and trucks. It contains 6000 images of each class.
The ensemble consists of $55$ base learners, $10$ of which specialize on $1$ class and $ {10 \choose 2} = 45$ of which specialize on $2$ classes.

\paragraph{Age estimation.}
UTKFace is a face images dataset consisting of over 20000 samples and different version of images format. Here 9700 cropped and aligned images are split in $5$ classes: baby (up to 5 years old), teenager (from $6$ to $19$), young (from $20$ to $33$), adult (from $34$ to $56$) and senior (more than $56$ years old). The classes are not uniformly distributed per number of ages, but each class contains the same number of samples. The goal is to estimate the person age given the face image.
The ensemble consists of $15$ base learners, 5 of which specialize on $1$ class and ${5 \choose 2} = 10$ on $2$ classes.

\paragraph{Emotion detection.}
Fer2013 is a dataset of over 30000 48x48 pixel grayscale face images, which are grouped in 7 classes: angry, disgust, fear, happy, neutral, sad and surprises. The goal is to categorize the emotion shown in the facial expression into one category.
 The ensemble consists of $21$ base learners, $7$ of which specialize on $1$ class and $ {7 \choose 2} = 21$ of which specialize on $2$ classes.

\begin{table}[tb]
\centering
\resizebox{0.85\linewidth}{!}
{
    \begin{tabular}{r ccccc}
    \toprule
        & \multicolumn{4}{c}{\textbf{Accuracy (\%)}}\\
         \textbf{Dataset} & \textbf{e2e-CEL}& \textbf{SL} & \textbf{UA} & \textbf{PV}  & \textbf{RS}\\
  \midrule
  {\textbf{MNIST}}   & \bf{98.55} & 96.88 & 96.81 & 95.99 & 96.83 \\ 
  {\textbf{UTKFACE}} & \bf{90.97} & 85.07 & 84.60 & 80.78 & 84.60 \\ 
  {\textbf{FER2013}} & \bf{66.31} & 64.95 & 63.89 & 63.15 & 63.89 \\ 
  {\textbf{CIFAR10}} & \bf{64.09} & 60.13 & 60.59 & 60.35 & 60.59 \\ 

  \bottomrule
\end{tabular}
}
  \caption{e2e-CEL vs super learner (SL), unweighted average (UA), 
  plurality voting (MV), and random selection (RS), 
  using specialized base learners.}
\label{tab:overall_results}
\end{table}

\subsection{e2e-CEL Analysis}
\label{sec:Results}
The \textit{e2e-CEL} strategy is tested on each experimental task for sub-ensemble size $k$ varying between $1$ and $n$, and compared to the baseline methods described above. Note in each case that accuracy is defined as the percentage of correctly classified samples over the master test set. 

\begin{figure*}[t]
    \centering
    \includegraphics[width=0.24\linewidth]{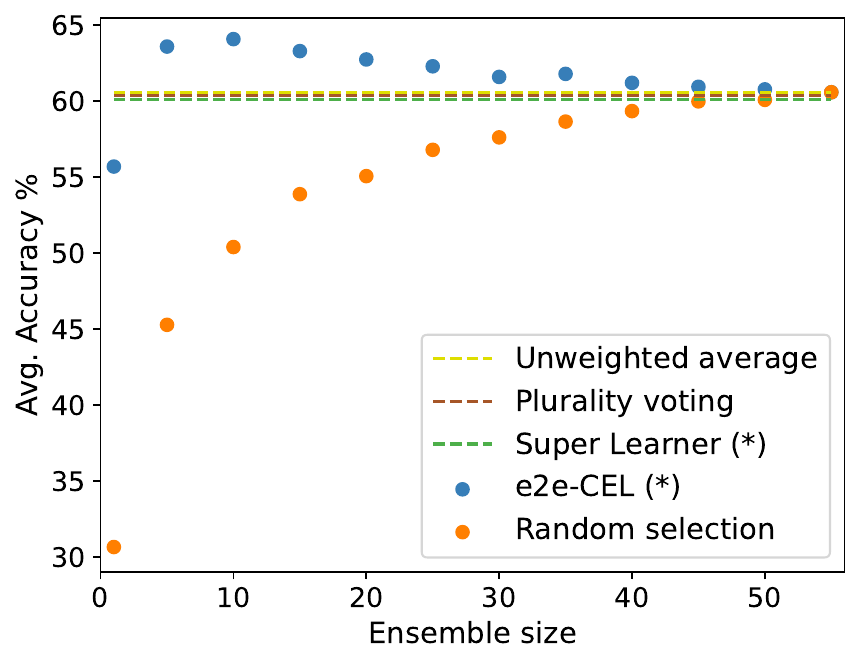}
    \includegraphics[width=0.24\linewidth]{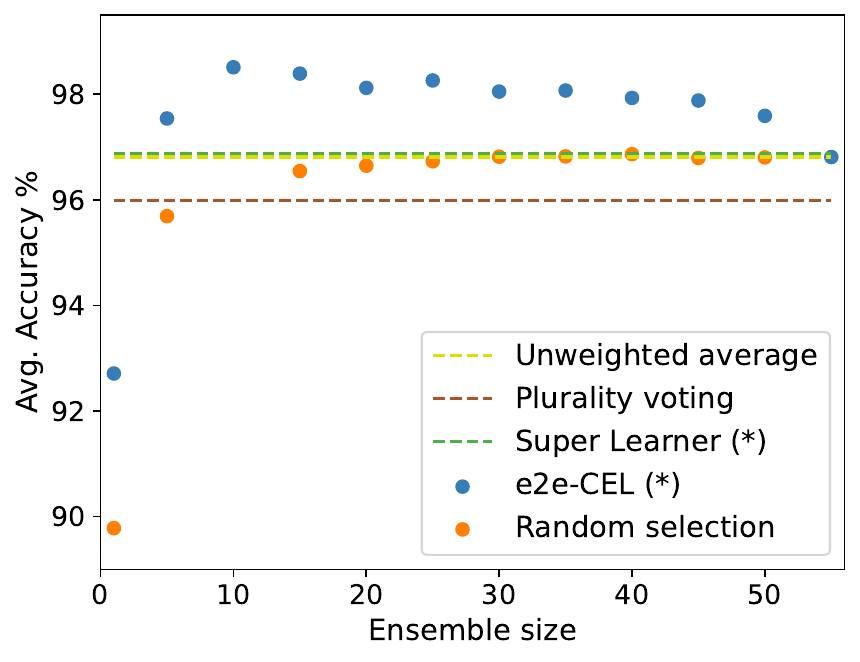} 
    \includegraphics[width=0.24\linewidth]{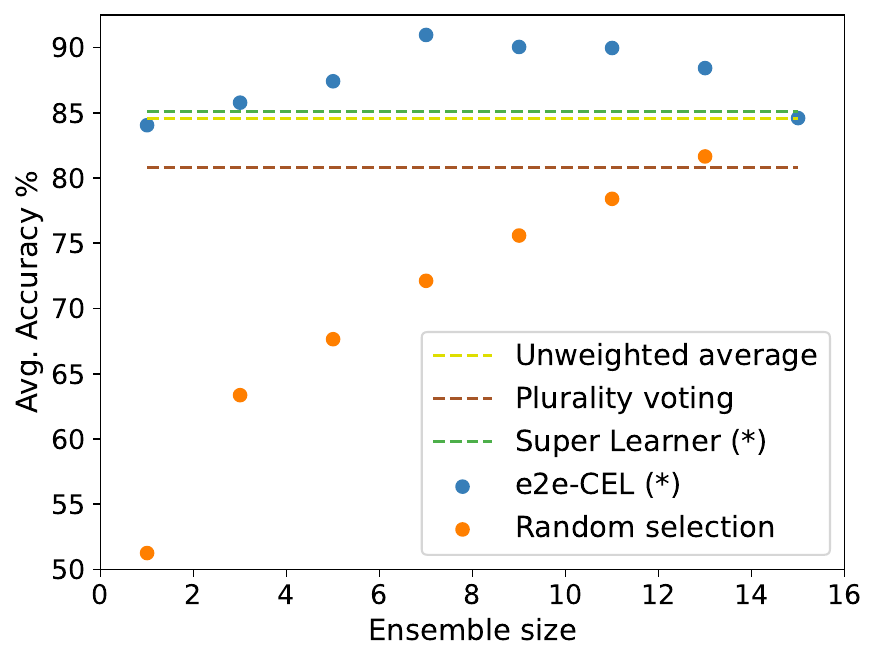}
    \includegraphics[width=0.24\linewidth]{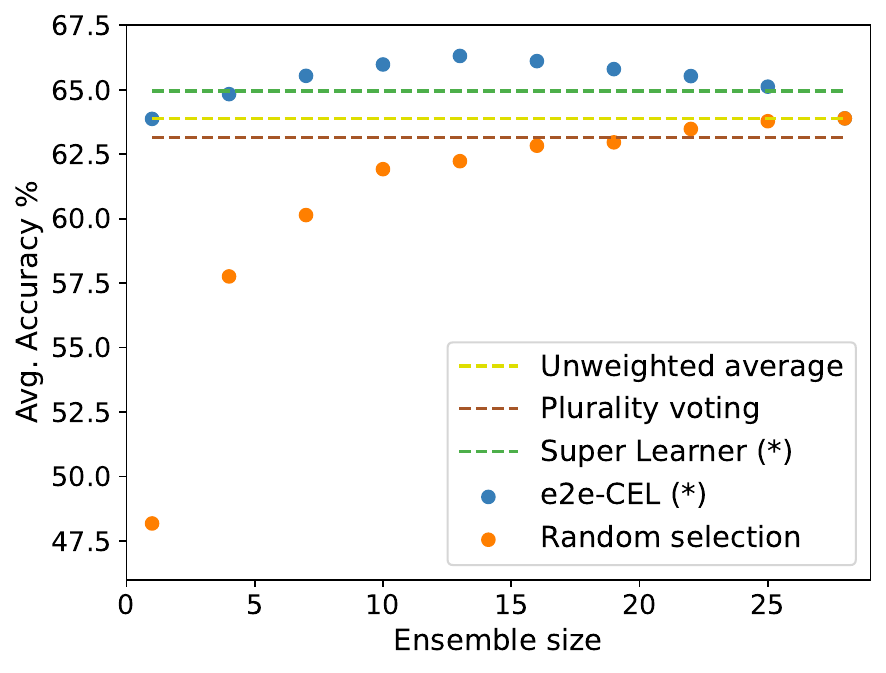}
\caption{Comparison between e2e-CEL and other ensemble 
models at varying of the sub-ensemble size $k$  on 
image classification--CIFAR10--(top left), 
digit classification--MNIST--(top right),
age estimation--UTKFace--(bottom left), and
emotion detection--FER2013--(bottom right) {The (*) in the label identifies methods that use specialized aggregation rules for every input sample.}.
\label{fig:2}}
\end{figure*}

Table \ref{tab:overall_results} reports the best accuracy over all the ensemble sizes $k$ of ensembles trained by e2e-CEL along with that of each baseline ensemble model, where each are formed using the same pre-trained base learners. Note how the proposed e2e-CEL scheme outperforms all the baseline methods, in each task, for all but the lowest values of $k$. 

Figure \ref{fig:2} reports the test accuracy found by e2e-CEL and ensembles based on the Super Learner, weighted average, majority voting, or random selection scheme. We make two key  observations:
{\bf (1)} Note from each subplot in Figure \ref{fig:2} that smart ensembles of size $k>1$ provide more accurate predictions than baseline models that randomly select sub-ensembles of the same size, a trend that diminishes as $k$ increases and base learner selections have less consequence (the two perform equally when $k=n$).
{\bf (2)} In every case, the sub-ensemble size which results in optimal performance is strictly between $1$ and $n$. 
{\em Importantly, this illustrates the motivating intuition of the e2e-CEL ensemble training. Neither the full ensemble ($k=n$), nor smart selection of a single base learner model ($k=1$) can outperform models that use smart selection of a sub-ensemble of any size.} 
A well-selected sub-ensemble  has higher potential accuracy than the master ensemble, and is, on average, more reliable than a well-selected single base learner.

\begin{table}[th]
\centering
\resizebox{\linewidth}{!}
{
    \begin{tabular}{r cc | cc}
      \toprule
  & & & \multicolumn{2}{c}{\textbf{Accuracy (\%)~~~~}}\\
 \textbf{Dataset} & \textbf{Classes}& \textbf{Best $k$} & 
 \textbf{e2e-CEL}  & \textbf{Individual base learners} \\
      \midrule
  \textbf{MNIST} & 10 & 10 &   \textbf{98.55} & 89.6\\ 
  \textbf{UTKFACE} & 5 & 7 &   \textbf{90.97} & 51.2\\ 
  \textbf{FER2013} & 7 & 13 &  \textbf{66.31} & 47.8\\ 
  \textbf{CIFAR10} & 10 & 10 & \textbf{64.09} & 31.1\\ 
  \bottomrule
    \end{tabular}
}

\caption{Left: Best ensemble size (Best $k$) and associated e2e-CEL test accuracy attained on each dataset. 
Right: Average accuracy for the constituent ensemble base learners.
\label{tab:e2e-CELresults}}
\end{table}

Next, Table \ref{tab:e2e-CELresults} (left) reports the accuracy of the e2e-CEL model trained on each task, along with the sub-ensemble size that resulted in highest accuracy. In two cases, for the digit classification and the image classification task, the e2e-CEL performs best when the sub-ensemble size is equal to the number of classes. In the remaining tasks, this observation holds approximately. {\em This is intuitive, since the number of base learners specializing on any class is equal to the number of classes, and e2e-CEL is able to increase ensemble accuracy by learning to select these base learners for prediction.}

Finally, observe the accuracy of e2e-CEL in Table \ref{tab:e2e-CELresults} (left) and the performance of the individual base learners predictors of the ensemble tested on both the labels in which their training was specialized as well as the other labels. Note how e2e-CEL predictions outperform their constituent base learners by a wide margin on each task. For example, on UTKFace, the e2e-CEL ensemble reaches an accuracy $40$ percentage points higher than its average constituent base learner. 
{\em This illustrates the ability of e2e-CEL to leverage the error diversity of base learners to form accurate classifiers by composing them based on input features}, even when the individual base learner's accuracies are poor.

  
    

\section{Conclusion}

This paper was motivated by the desire to address significant challenges in model selection and ensemble learning: the identification of the optimal model or ensemble for classifying any particular input sample. The proposed solution is a novel framework for differentiable model selection tailored for ensemble learning, integrating machine learning and combinatorial optimization. This framework is motivated by the idea that well-selected sub-ensembles can form more accurate predictions than their original ensemble. It adaptively selects sub-ensembles according to individual input samples.

The paper shows how to derive the ensemble learning task into a differentiable selection program which is trained end-to-end within the ensemble learning model. This approach allows the proposed framework to compose accurate classification models even from ensemble base learners with low accuracy, a feature not shared by existing ensemble learning approaches. The results on various tasks demonstrate the versatility and effectiveness of the proposed framework, substantially outperforming  state-of-the-art and conventional consensus algorithms in a variety of settings.

This work demonstrates that the integration of machine learning and combinatorial optimization is a valuable toolset for not only enhancing but also combining machine learning models to improve performance on common tasks. This work hopes to motivate new solutions where decision-focused learning may be used to improve the capabilities of machine learning systems. The proposed framework contributes to the ongoing efforts to improve the efficiency and effectiveness of model selection in machine learning, particularly in the context of ensemble learning.




\section*{Ethics Statement}
While the proposed e2e-CEL method has the potential to improve the performance of ensemble learning, it is important to consider the ethical implications of its use and to take steps to mitigate any potential negative impacts. 
One possible concern is the potential to select sub-ensembles in a way that could perpetuate or amplify biases present in the ensemble base learners. This could lead to unfair or discriminatory predictions for certain groups of people. 

It is also important to consider the potential benefit of this study. This approach allows for the composition of accurate classification models even from ensemble base learners with low accuracy or strong biases, which is a feature not shared by existing ensemble learning approaches. The proposed solution thus aims to enhance the performance of ensemble learning models and may advance the development of more effective predictors which exhibit fewer biases.

\section*{Acknowledgements}
This research is partially supported by NSF grants 2007164, 2232054, and NSF CAREER Award 2143706. Fioretto is also supported by an Amazon Research Award and a Google Research Scholar Award. Its views and conclusions are those of the authors only.

\section*{Contribution Statement}
James Kotary and Vincenzo Di Vito are both considered first authors. All authors have equal contributions.

\bibliographystyle{named}
\bibliography{ijcai23}

\begin{thebibliography}{}

\bibitem[\protect\citeauthoryear{Abernethy \bgroup \em et al.\egroup
  }{2016}]{abernethy2016perturbation}
Jacob Abernethy, Chansoo Lee, and Ambuj Tewari.
\newblock Perturbation techniques in online learning and optimization.
\newblock {\em Perturbations, Optimization, and Statistics}, 233, 2016.

\bibitem[\protect\citeauthoryear{Bazaraa \bgroup \em et al.\egroup
  }{2008}]{bazaraa2008linear}
Mokhtar~S Bazaraa, John~J Jarvis, and Hanis~D Sherali.
\newblock {\em Linear programming and network flows}.
\newblock John Wiley \& Sons, 2008.

\bibitem[\protect\citeauthoryear{Berthet \bgroup \em et al.\egroup
  }{2020}]{berthet2020learning}
Quentin Berthet, Mathieu Blondel, Olivier Teboul, Marco Cuturi, Jean-Philippe
  Vert, and Francis Bach.
\newblock Learning with differentiable pertubed optimizers.
\newblock {\em Advances in neural information processing systems},
  33:9508--9519, 2020.

\bibitem[\protect\citeauthoryear{Cormen \bgroup \em et al.\egroup
  }{2022}]{cormen2022introduction}
Thomas~H Cormen, Charles~E Leiserson, Ronald~L Rivest, and Clifford Stein.
\newblock {\em Introduction to algorithms}.
\newblock MIT press, 2022.

\bibitem[\protect\citeauthoryear{Deng}{2012}]{mnistDataset}
Li~Deng.
\newblock The mnist database of handwritten digit images for machine learning
  research [best of the web].
\newblock {\em IEEE Signal Processing Magazine}, 29(6):141--142, 2012.

\bibitem[\protect\citeauthoryear{Diederik and Jimmy}{2014}]{diederik2014adam}
PK~Diederik and B~Jimmy.
\newblock Adam: A method for stochastic optimization. iclr.
\newblock {\em arXiv preprint arXiv:1412.6980}, 2014.

\bibitem[\protect\citeauthoryear{Dona and
  Gallinari}{2021}]{dona2021differentiable}
J{\'e}r{\'e}mie Dona and Patrick Gallinari.
\newblock Differentiable feature selection, a reparameterization approach.
\newblock In {\em Machine Learning and Knowledge Discovery in Databases.
  Research Track: European Conference, ECML PKDD 2021, Bilbao, Spain, September
  13--17, 2021, Proceedings, Part III 21}, pages 414--429. Springer, 2021.

\bibitem[\protect\citeauthoryear{Folland}{1999}]{folland1999real}
Gerald~B Folland.
\newblock {\em Real analysis: modern techniques and their applications},
  volume~40.
\newblock John Wiley \& Sons, 1999.

\bibitem[\protect\citeauthoryear{Fu \bgroup \em et al.\egroup
  }{2016}]{fu2016drmad}
Jie Fu, Hongyin Luo, Jiashi Feng, Kian~Hsiang Low, and Tat-Seng Chua.
\newblock Drmad: Distilling reverse-mode automatic differentiation for
  optimizing hyperparameters of deep neural networks, 2016.

\bibitem[\protect\citeauthoryear{Gumbel}{1954}]{gumbel1954statistical}
Emil~Julius Gumbel.
\newblock {\em Statistical theory of extreme values and some practical
  applications: a series of lectures}, volume~33.
\newblock US Government Printing Office, 1954.

\bibitem[\protect\citeauthoryear{Ju \bgroup \em et al.\egroup
  }{2018}]{ju2018relative}
Cheng Ju, Aur{\'e}lien Bibaut, and Mark van~der Laan.
\newblock The relative performance of ensemble methods with deep convolutional
  neural networks for image classification.
\newblock {\em Journal of Applied Statistics}, 45(15):2800--2818, 2018.

\bibitem[\protect\citeauthoryear{Kairouz \bgroup \em et al.\egroup
  }{2021}]{federatedLearning}
Peter Kairouz, H.~Brendan McMahan, Brendan Avent, Aurélien Bellet, Mehdi
  Bennis, Arjun~Nitin Bhagoji, Kallista Bonawitz, Zachary Charles, Graham
  Cormode, Rachel Cummings, Rafael G.~L. D’Oliveira, Hubert Eichner, Salim~El
  Rouayheb, David Evans, Josh Gardner, Zachary Garrett, Adrià Gascón, Badih
  Ghazi, Phillip~B. Gibbons, Marco Gruteser, Zaid Harchaoui, Chaoyang He, Lie
  He, Zhouyuan Huo, Ben Hutchinson, Justin Hsu, Martin Jaggi, Tara Javidi,
  Gauri Joshi, Mikhail Khodak, Jakub Konecný, Aleksandra Korolova, Farinaz
  Koushanfar, Sanmi Koyejo, Tancrède Lepoint, Yang Liu, Prateek Mittal,
  Mehryar Mohri, Richard Nock, Ayfer Özgür, Rasmus Pagh, Hang Qi, Daniel
  Ramage, Ramesh Raskar, Mariana Raykova, Dawn Song, Weikang Song, Sebastian~U.
  Stich, Ziteng Sun, Ananda~Theertha Suresh, Florian Tramèr, Praneeth
  Vepakomma, Jianyu Wang, Li~Xiong, Zheng Xu, Qiang Yang, Felix~X. Yu, Han Yu,
  and Sen Zhao.
\newblock Advances and open problems in federated learning.
\newblock {\em Foundations and Trends® in Machine Learning}, 14(1–2):1--210,
  2021.

\bibitem[\protect\citeauthoryear{Kotary \bgroup \em et al.\egroup
  }{2022}]{kotary2022endtoend}
James Kotary, Vincenzo~Di Vito, and Ferdinando Fioretto.
\newblock End-to-end optimization and learning for multiagent ensembles, 2022.

\bibitem[\protect\citeauthoryear{Krizhevsky}{2009}]{cifarDataset}
Alex Krizhevsky.
\newblock Learning multiple layers of features from tiny images.
\newblock 2009.

\bibitem[\protect\citeauthoryear{Liu \bgroup \em et al.\egroup
  }{2016}]{ferDataset}
Kuang Liu, Mingmin Zhang, and Zhigeng Pan.
\newblock Facial expression recognition with cnn ensemble.
\newblock In {\em 2016 International Conference on Cyberworlds (CW)}, pages
  163--166, 2016.

\bibitem[\protect\citeauthoryear{Mienye and Sun}{2022}]{mienye2022survey}
Ibomoiye~Domor Mienye and Yanxia Sun.
\newblock A survey of ensemble learning: Concepts, algorithms, applications,
  and prospects.
\newblock {\em IEEE Access}, 10:99129--99149, 2022.

\bibitem[\protect\citeauthoryear{Paszke \bgroup \em et al.\egroup
  }{2019}]{NEURIPS2019_9015}
Adam Paszke, Sam Gross, Francisco Massa, Adam Lerer, James Bradbury, Gregory
  Chanan, Trevor Killeen, Zeming Lin, Natalia Gimelshein, Luca Antiga, Alban
  Desmaison, Andreas Kopf, Edward Yang, Zachary DeVito, Martin Raison, Alykhan
  Tejani, Sasank Chilamkurthy, Benoit Steiner, Lu~Fang, Junjie Bai, and Soumith
  Chintala.
\newblock Pytorch: An imperative style, high-performance deep learning library.
\newblock In {\em Advances in Neural Information Processing Systems 32}, pages
  8024--8035. Curran Associates, Inc., 2019.

\bibitem[\protect\citeauthoryear{Ribeiro \bgroup \em et al.\egroup
  }{2015}]{MLAS}
Mauro Ribeiro, Katarina Grolinger, and Miriam~A.M. Capretz.
\newblock Mlaas: Machine learning as a service.
\newblock In {\em 2015 IEEE 14th International Conference on Machine Learning
  and Applications (ICMLA)}, pages 896--902, 2015.

\bibitem[\protect\citeauthoryear{Ruder}{2016}]{ruder2016overview}
Sebastian Ruder.
\newblock An overview of gradient descent optimization algorithms.
\newblock {\em arXiv preprint arXiv:1609.04747}, 2016.

\bibitem[\protect\citeauthoryear{Sheth and
  Fusi}{2020}]{sheth2020differentiable}
Rishit Sheth and Nicol{\'o} Fusi.
\newblock Differentiable feature selection by discrete relaxation.
\newblock In {\em International Conference on Artificial Intelligence and
  Statistics}, pages 1564--1572. PMLR, 2020.

\bibitem[\protect\citeauthoryear{Wilder \bgroup \em et al.\egroup
  }{2019}]{wilder2019melding}
Bryan Wilder, Bistra Dilkina, and Milind Tambe.
\newblock Melding the data-decisions pipeline: Decision-focused learning for
  combinatorial optimization.
\newblock In {\em AAAI}, volume~33, pages 1658--1665, 2019.

\bibitem[\protect\citeauthoryear{Witten \bgroup \em et al.\egroup
  }{2005}]{witten2005practical}
Ian~H Witten, Eibe Frank, Mark~A Hall, Christopher~J Pal, and MINING DATA.
\newblock Practical machine learning tools and techniques.
\newblock In {\em Data Mining}, volume~2, 2005.

\bibitem[\protect\citeauthoryear{Zeiler}{2012}]{zeiler2012adadelta}
Matthew~D Zeiler.
\newblock Adadelta: an adaptive learning rate method.
\newblock {\em arXiv preprint arXiv:1212.5701}, 2012.

\bibitem[\protect\citeauthoryear{Zhifei~Zhang}{2017}]{utkfaceDataset}
Hairong~Qi Zhifei~Zhang, Yang~Song.
\newblock Age progression/regression by conditional adversarial autoencoder.
\newblock In {\em IEEE Conference on Computer Vision and Pattern Recognition
  (CVPR)}. IEEE, 2017.

\end{thebibliography}

\appendix \section{Related Work}
\label{app:rel_work}

This paper proposes a framework for composing effective ensemble learning models by adapting techniques from the intersection of constrained optimization and machine learning. This section briefly reviews the two as-yet distinct areas:

\paragraph{Ensemble learning.}
Several works have focused on studying ensemble models for more effective machine learning. We refer the reader to  \cite{DBLP:journals/fcsc/DongYCSM20} for a useful categorization of ensemble methods, depending on the underlying machine learning task to be solved. 

Ensemble learning often involves two distinct aspects: {\bf (1)} the training of individual ensemble learning base learners, and {\bf (2)} the aggregation of their individual outputs into accurate ensemble predictions. The former aspect concerns the composition of an ensemble from base learners with complementary error profiles, and is commonly handled by \emph{bagging}, which consists of randomly composing datasets for training each ensemble member,  and \emph{boosting}, which involves adaptively contriving a sequence of datasets based on the error distributions of their resulting models so that error diversity is increased. Many variations and task-specific alternatives have also been proposed \cite{DBLP:journals/fcsc/DongYCSM20}. For example, \cite{ensemble_pruning} studies the effects of various pruning techniques for reducing the ensemble size from an initial bagging ensemble. A survey focusing on the training of individual ensemble base learners is provided in \cite{mienye2022survey}.

The latter aspect is typically handled by classical aggregation rules over either the discrete class predictions or the continuous activation values of ensemble members. \emph{Majority} and \emph{plurality voting} are consensus criteria for choosing a discrete class prediction based on the most popular among ensemble members. Some works have attempted mathematically model more effective aggregation rules: these include \emph{super-learners} \cite{ju2018relative}, which attempt to form a weighted combination of base learner models that maximizes accuracy over a validation set. 

\paragraph{Decision focused learning.}
A body of work has been devoted to studying constrained optimization problems as learnable components within neural networks. See for example the recent survey of \cite{kotary2021end}. As a mapping between some problem-defining parameters and their optimal solutions, an optimization problem can be viewed as a function whose outputs adhere to explicitly defined constraints. In this way, special structure can be guaranteed within the predictions \cite{elmachtoub2020smart} or learned embeddings of a neural network \cite{amos2019optnet}. The primary challenge is that differentiation of the optimization mapping is required for backpropagation. Distinct approaches have been proposed, which depend on various approximations, depending on the optimization problem's form \cite{elmachtoub2020smart}, \cite{vlastelica2020differentiation}, and  \cite{ferbermipaal}. Smooth convex problems, in particular, admit exact gradients via their well-defined equations for optimality, as shown by \cite{amos2019optnet} and \cite{agrawal2019differentiable}. Convex problems which define piecewise-constant mappings, such as  linear programs, require approximation by smooth problems before differentiation. Smoothing techniques are commonly applied to the objective function and include the addition of regularizing objective terms, as proposed by \cite{wilder2019melding}, and random noise, as in  \cite{berthet2020learning}. 

In this paper, the desired structured prediction primarily takes the form of a binary vector representing subset selection, in which $k<n$ among $n$ items are indicated for selection. This work exploits the fact  that differentiation of this structure as a linear program with respect to a parameterizing vector allows it to be applied in an end-to-end differentiable masking scheme that is used to select and combine the best ensemble base learners' predictions for a given input sample.

\begin{table}[!t]
    \centering
    \resizebox{0.95\linewidth}{!}{
     \begin{tabular}{l l} 
     \toprule
     \textbf{Symbol} & \textbf{Semantic}\\
     \midrule
     $n$      & Size of the ensemble\\
     $k$      & Size of a sub-ensemble\\
     $m$      & Number of noise samples for perturbed optimizer\\
     $f_i$    & Pre-trained classifier, $1 \leq i \leq n$\\
     $\hat{y}_i$    & Softmax prediction from $f_i$\\
     $f$      & Smart ensemble classifier\\
     $\hat{y}$    & Smart ensemble softmax prediction \\
     $g$     & Selection net\\
     $\theta$     & Selection net parameters\\
     $\hat{c}$    & Predicted scores from the selection net \\
     $\mathcal{K}$   & Unweighted knapsack function\\
     $\mathbb{K}$   & Perturbed unweighted knapsack function\\
     $\mathcal{X}$    & Input feature data \\
     $\mathcal{Y}$    & Target class data \\
     $\mathcal{L}$    & Classification loss function \\
     \bottomrule
    \end{tabular}  
}
\caption{Common symbols}
\label{table:symbols}
\end{table}

\section{Datasets and settings}
\label{dataSet} 
This section provides information about the base learner's model architecture and the training dataset composition for each task.

\paragraph{Digit classification.}
Here the base learners use convolutional neural networks (CNN), each of those with the same architecture: $2$ convolutional layers and $3$ linear layers. Within the convolutional layers, of $10$ and $20$ output channels respectively, the information is processed using a square kernel filter of size $5$x$5$. The training dataset of each base learner is composed, on average, by $73.2\%$ of samples belonging to one (or two) specific class(es), while the remaining part is uniformly distributed over all the other classes by random sampling. On average, this results in $97.5\%$ of accuracy in classifying a particular subset of training samples and $86.8\%$ of accuracy in classifying the complementary subset of training samples.\\\textbf{Image classification} \\
Here the base learners use CNN models, each of those with the same architecture, $2$ convolutional layer, each containing a square kernel filter of size $5$, with $6$ and $16$ output channels respectively, and $3$ linear layers.
On average, the training dataset of each base learner is composed by $55.2\%$ of samples belonging to one (or two) specific class(es), while the remaining part is uniformly distributed over all the other classes by random sampling. This results in $76.3\%$ accuracy for classifying a particular subset of training samples and $24.8\%$ accuracy for classifying the complementary subset of training samples. 

\paragraph{Age estimation.}
Here the base learner model architecture is a pretrained version of Resnet18 on the ImageNet-1k dataset, which is entirely re-trained.
On average, the training dataset of each base learner is composed by $58.3\%$ of samples belonging to one (or two) specific class(es), while the remaining part is uniformly distributed over all the other classes by random sampling. This results in $93.2\%$ accuracy for classifying a particular subset of training samples and $25.2\%$ accuracy for classifying the complementary subset of training samples.

\paragraph{Emotion detection.}
Here the base learner model architecture consists of $8$ convolutional layer and a linear layer. For each convolutional layer, the number of output channels is $64, 128, 128, 128, 256, 512, 512$ and $512$; each uses a square kernel of size 3x3.
The training dataset of each base learner is composed by $44.4\%$ of samples belonging to one (or two) specific class(es), while the remaining part is uniformly distributed over all the other classes by random sampling. This results in $79.4\%$ accuracy for classifying a particular subset of training samples and $38.1\%$ accuracy for classifying the complementary subset of training samples.

\section{Hyperparameters}

\subsection{Hyperparameters of the base learners' model training}

Here for each task, additional details about the hyperparameters of the base learners' model training are provided .

\begin{table}[h]

\centering
\resizebox{\linewidth}{!}
{
    \begin{tabular}{r cccccc} 
      \toprule
         \textbf{Dataset}& \textbf{Optimizer}& \textbf{Scheduler} & \textbf{Learning rate} &  \textbf{Gamma} & \textbf{Epochs} & \textbf{Loss function} \\
      \midrule
      {\textbf{MNIST}} & Adadelta & StepLR & 1 & 0.9 & 14 & Cross entropy\\ 
      {\textbf{UTKFACE}} & Adadelta & StepLR & 1 & 0.9 & 10 & Cross entropy\\
      {\textbf{FER2013}} & Adam & OneCycleLR &  0.001  & & 70 & Cross entropy\\ 
      {\textbf{CIFAR10}} & SGD & & 0.001 & &  3 & Cross entropy\\ 
      \bottomrule
    \end{tabular}
}
\label{table:base_learner_hyperparams}
\end{table} 

\subsection{Hyperparameters of the selection net's model training}

Here, for each task, additional details about the hyperparameters of the selection net's model training are provided. We recall that for each task, the selection net's model share the same architecture of the corresponding base learner's model.

\begin{table}[h]

\centering
\resizebox{\linewidth}{!}
{
    \begin{tabular}{r ccccccc} 
      \toprule
         & \textbf{Optimizer}& \textbf{Scheduler} & \textbf{Learning rate} & \textbf{Gamma} & \textbf{Epochs} & \textbf{Loss function} \\
      \midrule
      {\textbf{MNIST}} & Adadelta & & 1 & & 8 & Cross entropy\\ 
      {\textbf{UTKFACE}} & Adadelta & & 1 &  & 10 & Cross entropy\\
      {\textbf{FER2013}} & Adam & OneCycleLR &  0.001  & & 20 & Cross entropy\\ 
      {\textbf{CIFAR10}} & Adadelta & & 0.001 & & 3 & Cross entropy\\ 
      \bottomrule
    \end{tabular}
}
\end{table}

\end{document}